\documentclass[pdflatex,sn-nature]{sn-jnl}

\usepackage{graphicx}%
\usepackage{multirow}%
\usepackage{amsmath,amssymb,amsfonts}%
\usepackage{amsthm}%
\usepackage{mathrsfs}%
\usepackage[title]{appendix}%
\usepackage{xcolor}%
\usepackage{textcomp}%
\usepackage{manyfoot}%
\usepackage{booktabs}%
\usepackage{algorithm}%
\usepackage{algorithmicx}%
\usepackage{algpseudocode}%
\usepackage{listings}%

\theoremstyle{thmstyleone}%
\theoremstyle{thmstyletwo}%

\theoremstyle{thmstylethree}%

\begin{document}
\title[The taggedPBC]{The \emph{taggedPBC}: Annotating a massive parallel corpus for crosslinguistic
investigations}
\author{\fnm{Hiram} \sur{Ring}}\email{hiram.ring@ntu.edu.sg}


\affil{
\orgname{Nanyang Technological University}, 
\orgaddress{
\country{Singapore}}}
\abstract{Existing datasets available for crosslinguistic investigations have
tended to focus on large amounts of data for a small group of languages
or a small amount of data for a large number of languages. This means
that claims based on these datasets are limited in what they reveal
about universal properties of the human language faculty. While this
has begun to change through the efforts of projects seeking to develop
tagged corpora for a large number of languages, such efforts are still
constrained by limits on resources. The current paper reports on a
large tagged parallel dataset which has been developed to partially
address this issue. The \emph{taggedPBC} contains POS-tagged parallel
text data from more than 1,940 languages, representing 155 language
families and 78 isolates, dwarfing previously available resources.
The accuracy of particular tags in this dataset is shown to correlate
well with existing SOTA taggers for high-resource languages (SpaCy,
Trankit). Additionally, a novel measure derived from this dataset,
the N1 ratio, correlates with expert determinations of intransitive
word order in three typological databases (WALS, Grambank, AUTOTYP)
such that a Gaussian Naive Bayes classifier trained on this feature
can accurately identify basic intransitive word order for languages
not in those databases. While much work is still needed to expand
and develop this dataset, the \emph{taggedPBC} is an important step
to enable corpus-based crosslinguistic investigations, and is made
available for research and collaboration via GitHub.\footnote{This version of the article has been accepted for publication, and is subject to Springer Nature’s AM terms of use, but is not the Version of Record and does not reflect post-acceptance improvements, or any corrections. The Version of Record is available online at: https://doi.org/10.1007/s10579-026-09902-2}}
\maketitle

\section{Introduction}

There are over 7,000 languages in the world, and such languages are
rather diverse in terms of their typological features. At the same
time, despite the overall diversity there are many similarities between
languages, suggesting the possibility of universal properties that
guide language evolution. These diversities and similarities have
been identified and categorized in various ways by a variety of databases,
most notably WALS \cite{Dryer:2013ab}, Grambank \cite{Skirgard:2023ab}, AUTOTYP \cite{Bickel:2023aa},
Glottolog \cite{Hammarstrom:2024aa}, and Lexibank \cite{List:2022aa}.\footnote{See http://wals.info ; http://grambank.clld.org ; http://github.com/autotyp/autotyp-data
; http://www.glottolog.org ; http://lexibank.clld.org}
There are also collections of corpora that are expansive in their
coverage of individual or groups of languages yet focused on particular
genres or domains (CHILDES, Brown, etc; and see the extensive corpora
available via the CLARIN project).\footnote{See http://childes.talkbank.org ; http://www.clarin.eu/resource-families}
The former tend to provide information \emph{about} languages, while
the latter tend to provide actual language \emph{data} in various
forms. The issue with the former is often lack of coverage for particular
features for all languages in the dataset, whereas the latter typically
suffer from insufficient annotation and lack linguistic coverage.
Both kinds of datasets have a particular kind of value, but do not
go far enough to address the concerns of linguistic diversity (i.e.
\cite{Benitez-Burraco:2025aa}) and gradience (i.e. \cite{Levshina:2023ab})
needed to expand our understanding of (cross-)linguistic processes.

More recent attempts to create annotated databases that do address
these issues by developing robust crosslinguistic corpora include
the Universal Dependencies Treebanks project (UDT; \cite{Zeman:2024ab}),\footnote{See http://universaldependencies.org}
which contains data from individual languages that has been annotated
for parts of speech and dependencies using a common framework. These
kinds of datasets are promising in allowing for the automatic identification
of properties that can categorize or classify languages in various
types, which reflects an increasing interest in computational typology
\cite{Jager:2025aa}, whereby annotated corpora can allow for
extraction of features that align with those found in typological
databases.\footnote{It should be noted here that so far the focus in computational typology
has been less on automatically extracting features from corpora and
more on validating theories of language/universals using expert classifications
found in various descriptive typological databases, a consideration
that I return to in the final discussion section below.\label{fn:It-should-be}} Unfortunately, such datasets still suffer many of the same issues
regarding coverage and degree of annotation as identified above.

The amount of crosslinguistic coverage and degree of annotation of
existing datasets are concerns when trying to investigate questions
related to universal properties of human language. The reason for
this is that such investigations require annotated data with information
on word types or dependencies, data unavailable for most of the world's
languages. This means that crosslinguistic investigations using smaller
datasets may suffer from lack of generalizability. For example, while
\cite{Hahn:2022aa} investigated dependencies in 80 languages (the
largest sample of languages to date for this kind of investigation),\footnote{Recent cross-linguistic investigations using tagged corpora include
\cite{Jing:2021aa} (71 languages) and \cite{Hahn:2020aa}
(51 languages).} the majority of languages in their dataset were from the Indo-European
language family. This calls into question the generalizability of
their findings, as such findings might be biased toward a particular
group of languages \cite{Blasi:2022aa}.\footnote{Similar to how other scientific findings have been biased toward WEIRD
demographics \cite{Henrich:2010aa}.}

In its most recent form, the UDT has 160+ hand-annotated languages
available for crosslinguistic investigation, but there are some concerns
with the data. For one thing, the datasets are not parallel texts,
which means there is a variety of genres representing each language,
such that one could question how representative they are of a particular
language as well as how easily they can be compared with corpora from
other languages. Additionally, the number of sentences available for
each language is variable, with many languages being represented by
fewer than 200 sentences in the v2.14 release. Other datasets of tagged
corpora have similar issues in terms of coverage. For example, the
ODIN project (and its re-packaged form via LECS)\footnote{See http://depts.washington.edu/uwcl/odin ; http://huggingface.co/datasets/lecslab/glosslm-corpus}
contains interlinear glossed text (IGT) from over 1,800 languages,
but the majority of these are single-word/line examples, and only
300 of the languages in their dataset have more than 100 sentences.

These datasets and related publications show the value of annotated
data for crosslinguistic investigation, but they also highlight some
of the difficulties inherent in creating such data. For one thing,
annotated data requires a large investment in time and manpower, whereby
experts in specific languages must invest time to do actual annotation
of the data. Additionally, annotators and researchers must decide
which features should be annotated and to what degree. This is also
partly dictated by the specifics of a language (which features are
relatively more rare/interesting and/or well-described) as well as
the amount, quality, and kind of descriptions available for a given
language. Generally the data is sourced with a particular goal or
research question in mind, which then conditions the manner and kind
of annotation that is done.

Recent advances in computational tools suggest the possibility of
using such tools for automatic annotation, greatly reducing the time
and effort needed to develop multilingual tagged datasets. As an example,
there is some indication that automated taggers for under-resourced
languages can benefit from high-resource language part of speech tags
in parallel corpora \cite{Agic:2015aa,Imani:2022aa}, particularly
for major word classes like nouns and verbs, even in zero-shot situations
on low-resource languages. But while the use of such tools may greatly
reduce the annotation burden for researchers, the derived dataset
must also be validated for crosslinguistic investigation. Specifically,
it is incumbent on researchers to identify a set of criteria used
to evaluate whether the automatic annotations are sufficiently aligned
with human-curated annotations or classifications before such a dataset
can be considered robust enough for further crosslinguistic investigation.

The current paper reports on the development of a large corpus of
parallel texts tagged for word class that partly addresses the issues
identified above. With data sourced from the Parallel Bible Corpus
(PBC; \cite{Mayer:2014aa}: 1,597 languages) and additional data
sourced from newly-available Bible translations (345 languages), computational
word alignment tools allow for automatic tagging of word classes in
1,942 languages (roughly 25\% of the world's languages) that represent
155 families and 78 isolates.\footnote{`Isolates' is defined here as languages found in the corpus that do
not share family membership with another language in the corpus.} The automated part of speech tagging method outlined in this paper
was validated by comparing the results with output from existing automatic
taggers for high-resource languages (SpaCy, Trankit).\footnote{See http://spacy.io/ ; http://nlp.uoregon.edu/trankit}
The results indicate good alignment, providing a baseline from which
to conduct crosslinguistic investigations. Further, features extracted
from this dataset align with word order classifications in typological
databases.

Statistical properties of the derived corpus also comport with expectations
regarding normality of language data. Importantly, this is partly
because each language has a large amount of data available for comparison.
Individual corpora in the majority of these languages (over 1,850)
contain more than 1,800 parallel sentences/verses, and the smallest
corpus contains more than 700 sentences/verses. This dwarfs both the
linguistic data and the language diversity available from other sources.
Further, the parallel nature of the data means that crosslinguistic
investigations using this dataset can be confident that they are comparing
similar contexts. The dataset (the \emph{taggedPBC}) as well as explanations
regarding its development and Python code for replicating the results
described here, can be found at the linked repository.\footnote{See http://github.com/lingdoc/taggedPBC\label{fn:linked-repo}}

In the sections below I first outline the methodology used to develop
the auto-taggers for each language (§\ref{sec:Methodology}). I then
highlight statistics of the dataset (§\ref{sec:Results}) and describe
the test used to assess tag validity (§\ref{subsec:Validating-tag-accuracy}).
I also show that this dataset allows for extraction of features associated
with typological properties, identifying a novel measure derived from
this dataset (the ``N1 ratio'') that correlates with intransitive
word order classifications in multiple typological databases (§\ref{subsec:Testing-features-extracted}).
I conclude the paper with discussion of ways the dataset could be
improved to support more robust crosslinguistic investigations (§\ref{subsec:Potential-for-expansion})
as well as potential avenues of research made possible by this dataset
(§\ref{subsec:Potential-research-directions}).

\section{Methodology\label{sec:Methodology}}

Transferring POS tags from high-resource languages has been shown
to work relatively well for word classes with many members, such as
Nouns and Verbs \cite{Agic:2015aa,Imani:2022aa,Imani:2022ab}.
Specifically, \cite{Agic:2015aa} show that using an IBM Model 2
to transfer tags between aligned words in pairs of languages achieves
much better results than unsupervised and weakly-supervised approaches.
Additionally, \cite{Imani:2022aa}, using the PBC, were able to improve
on accurate POS-tagging for low-resource languages by propagating
labels from 36 high-resource languages to 17 low-resource target languages.
They note that 
\begin{quote}
``A limitation of the GLP {[}our method{]} is that training over
a MAG (multilingual alignment graph) created for all PBC languages
requires a prohibitively large amount of resources, and based on our
experiments, if we use a larger number of target languages at the
same time, the performance will likely drop.'' (p. 1585)
\end{quote}
While \cite{Imani:2022aa}'s approach does result in higher overall
accuracy than that of \cite{Agic:2015aa} for specific language pairs,
it is not clear that it performs better for particular word classes
or for all low-resource languages more generally. Additionally, the
computational resources required by Imani et al's approach (using
transformer architectures and word embeddings) are much more significant
than those required by \cite{Agic:2015aa}'s approach. Accordingly,
the methodolgy used here follows that of \cite{Agic:2015aa} in training
IBM Model 2 to align words between language pairs, in this case each
source language and English. It should be noted here that IBM Model
2 is a statistical word alignment model that is agnostic to the order
of words, which makes it ideal for a situation like this one, where
we are dealing with languages for which we don't know the word order.
The essential idea is that for each language pair we train a model
to align words in a parallel corpus with two languages, then we ``translate''
the words in the source (low-resource) language to the target (high-resource)
language, and transfer the POS-tag of the target language word to
the source language word.

Another important thing to note is that while machine translation
methods have expanded in recent years, there are various issues with
some of the more recent (``AI''-related) approaches, particularly
the use of large language models. Training or fine-tuning such models
to translate between languages is not only computationally resource-intensive,
but often also result in inaccuracies in output, making it questionable
to transfer POS-tags. Further, developments in word alignment algorithms
such as \emph{eflomal} \cite{Ostling:2016aa} and \emph{Embedding-Enhanced
GIZA++} \cite{Marchisio:2022aa} require language-specific processing
(embeddings in some cases) and can also be computationally resource-intensive.
Additionally, alignment algorithms such as \emph{eflomal} do not handle
language pairs with word order differences particularly well. In contrast,
the IBM Model 2 has relatively minor computational resource requirements
and is simple to implement, while depending primarily on statistical
patterns that handle differences in word order between language pairs.

\subsection{Improving tag accuracy}

There are several additional steps that I take to improve tagging
accuracy, which fall under two general principles. The first principle
is related to ensuring that parallel verses share as many common concepts
as possible, i.e. that they are in some sense more semantically accessible.
Increasing the number of shared concepts between languages in a dataset
relative to rare concepts should allow a statistical word alignment
model to more successfully identify alignments between shared concepts
and/or words. This is akin to ``under-sampling'' in machine learning,
whereby relevant properties of a dataset are adjusted to reduce noise
in training. The second principle is related to ensuring that specific
properties of individual languages (with regards to word formation)
are taken into account. Since languages differ in terms of fusional
properties (how many morphemes or units of meaning a word contains),
this ought to be considered by any attempt at word alignment.

Accordingly, to address the concerns present under the first principle,
I reduce the number of verses in the corpus using the following criteria.
I first lemmatize two English New Testament translations, then reduce
the dataset to verses which share at least 4 lemmas in those translations.
This ensures that verses are more likely to be semantically accessible.
Translations chosen for this procedure are the New Living Translation
(NLT) and the New International Version (NIV), after discussions with
Bible translators who work with low-resource languages. Translators
indicated that these two versions are often consulted for their readability
and accessibility in terms of how they translate the original Greek
text, particularly in backtranslation (a checking step before final
translation verification). The smaller dataset is then further reduced
to verses where a verb occurs that is also present in at least five
other verses, which increases the likelihood that each individual
non-English Bible translation has common concepts that can be aligned
with words in the English text. This procedure results in 1,884 verses
for training word alignment models. As noted above, this is a systematic
method for under-sampling the parallel data, which should help improve
the quality of the trained statistical word alignment model.

In order to address the second principle, I train subword tokenizers
for each language. For this I leverage the MarianMT subword tokenizer,
which has been developed for machine translation between multiple
languages, as a base architecture. Specifically, I used the SentencePieceBPETokenizer
without lowercasing (from the Huggingface Transformers library) and
retrained tokenizers on each language. Training subword tokenizers
for a single language allows the tokenizer to identify sub-word units
that hypothetically correspond to morphemes. Using language-specific
tokenizers before mapping alignments between sentences should enhance
the ability of the statistical word alignment model to identify correspondences
between words in the pairs of languages.

It should also be noted that the languages in the PBC use several
different writing systems and scripts. Some, like Devanagari, are
syllabic, while others, like Arabic, are largely phonetic, and yet
others, like Chinese, are systems that use individual characters to
represent words or syllables. In order to ensure that we can adequately
compare languages in terms of their sound structure, and to support
the training of subword tokenizers, I convert all disparate systems
to romanized script using the \emph{uroman}\footnote{See http://github.com/isi-nlp/uroman}
tool, with the additional step of performing word-level tokenization
for translations using scripts where words are not clearly separated
by whitespace (\emph{fugashi}\footnote{See http://github.com/polm/fugashi}
for Japanese, \emph{khmernltk}\footnote{See http://github.com/VietHoang1512/khmer-nltk}
for Khmer/Kuy, \emph{jieba}\footnote{See http://github.com/fxsjy/jieba}
for Chinese, \emph{pyidaungsu}\footnote{See http://github.com/kaunghtetsan275/pyidaungsu}
for Burmese/Karen/Mon, \emph{attacut}\footnote{See http://github.com/PyThaiNLP/attacut}
for Thai, \emph{botok}\footnote{See http://github.com/OpenPecha/Botok}
for Tibetan) before converting to romanized script.\footnote{As noted by a reviewer, this is not an ideal mapping, since romanization
does not necessarily reflect the phonemic values of a language in
the same way that a standardized orthography does. Putting aside the
question of phonemic or phonetic accuracy, however, I would argue
that romanization does at least allow for comparison in a way that
retaining original scripts does not.}

As a final step, I train each model with the source being the non-English
language and the target being English. This enables a direct mapping
of the words in the source language verse to the English words, allowing
us to extract POS tags easily from words in the POS-tagged English
sentence (via the SpaCy library). That is, once the respective translation
model has been trained for a particular language, every word of every
sentence/verse is back-translated to English. If the translation of
that word matches a word in the respective POS-tagged English sentence/verse,
the source language word is tagged with the English POS.\footnote{For the purposes of cross-linguistic comparison, I employ the Universal
Dependencies tag set for parts of speech which includes 6 open classes
(ADJ ‘adjective’, ADV ‘adverb’, INTJ ‘interjection’, NOUN ‘noun’,
PROPN ‘proper noun’, VERB ‘verb’), 8 closed classes (ADP ‘adposition’,
AUX ‘auxiliary’, CCONJ ‘coordinating conjunction’, DET ‘determiner’,
NUM ‘numeral’, PART ‘particle’, PRON ‘pronoun’, SCONJ ‘subordinating
conjunction’) and 3 ‘other’ classes for handling characters (PUNCT
‘punctuation’, SYM ‘symbol’, X ‘other’). These 17 classes, while not
used by all languages, provide enough of a basic tag set to allow
for cross-linguistic comparison.\label{fn:For-the-purposes}} This methodology enables taggging of 1,884 verses for a large number
of languages in the PBC. Due to some differences in how sentences/clauses
are translated between languages, and some portions not being available
for some languages, there are a few mismatches. We can examine the
actual number of sentences and sort them into bins, which gives counts
of verses/languages in the corpus (Table \ref{tab:Number-of-verses}).

\begin{table}[h]
\noindent \begin{centering}
\begin{tabular}{lr}
\textbf{Number of verses} & \textbf{Number of languages}\tabularnewline
1800+ & 1853\tabularnewline
1500-1800 & 27\tabularnewline
1000-1500 & 50\tabularnewline
700-1000 & 12\tabularnewline
\textbf{Total} & \textbf{1,942}\tabularnewline
\end{tabular}
\par\end{centering}
\caption{Number of verses in each tagged language corpus\label{tab:Number-of-verses}}

\end{table}

\subsection{Caveats}

While automated transfer of POS tags from high-resource languages
such as English to low-resource languages has been shown to give decent
results, as noted by the literature above, there are some necessary
caveats. First, limiting the number of sentences does remove some
signal along with noise, meaning that it is possible not all word
types will be represented in a given language. Along with this, mapping
infrequent word types or function words (i.e. \emph{that} in English,
\emph{der} in German), which are often less prominent and more functionally
diverse crosslinguistically, will likely be less succcessful. Another
case in point is the presence of such terms as numeral classifiers:
many languages require special words or markers that classify nouns
being enumerated, but such a class of words is simply not present
in other languages, like English. Since we are transferring tags from
English, word classes like numeral classifiers will not be tagged
in this dataset at all, which means there is scope for improving on
these annotations. At the same time, the large number of parallel
sentences for the vast majority of languages in this dataset, and
the common method by which they were developed, means that findings
for other word types (specifically nouns, proper nouns, pronouns,
verbs, and auxiliaries) are statistically robust.

The methodology followed here should not be construed as a claim that
all languages share the same tag set, but rather that certain word
classes (nouns, verbs, etc.) exist in the majority of languages and
can be readily identified via translation. At the same time, there
are a large number of words in each of these corpora that simply have
no tag (more than 50\% of tokens in most corpora are tagged as ``unk'')
because there is no direct translation equivalent corresponding to
a tagged English word in the same verse. This reflects the methodological
goal of maximizing accuracy of POS transfer. While this leaves potentially
ambiguous words for future investigation (and see the discussion in
the concluding section below regarding development of language-specific
automated tools or manual annotation for individual low-resource corpora),
the translation method provides a decent proxy measure that supports
the identification of major word classes. As I demonstrate below,
despite the caveats presented here, information and features extracted
from this tagged dataset comport with expected observations and lead
to novel findings. This dataset is therefore released in order to
provide a baseline for future research, with potential for expansion
and refinement, particularly in relation to language-specific POS
tags.

\section{Results\label{sec:Results}}

The final \emph{taggedPBC} dataset contains tagged data for 1,942
modern languages (1,944 if you include the 2 conlangs with NT translations:
Esperanto and Klingon). This represents roughly 25\% of the world's
languages from 155 distinct families and 78 single languages or isolates.
The number of languages with more than 1,800 sentences/verses in the
dataset is 1,853 (see Table \ref{tab:Number-of-verses}), and the
fewest number of verses is represented by Nafusi (ISO 639-3: jbn)
with 729 verses. The automatic tagging method outlined above results
in a statistically robust number of unique arguments and predicates\footnote{In the current paper, I use the term ``argument'' and ``predicate''
as a way of grouping classes of words that loosely correspond to heads
of phrases and serve similar roles/functions crosslinguistically.
Specifically, ``argument'' groups the word classes of nouns, proper
nouns, and pronouns, which are deemed to represent prototypical agents/patients.
The term ``predicate'' groups the word classes of verbs and auxiliaries,
which are deemed to represent prototypical actions/states. By using
these terms to group POS tags, I am not claiming that these tags exclusively
represent all instantiations of “arguments” or “predicates” crosslinguistically,
but simply separating out those tags from other tags in the Universal
Tag Set used in the overall dataset, in order to develop procedures
to compute relevant statistics for crosslinguistic comparison.} for all 1,942 languages (mean: 699 arguments, 466 predicates), with
the fewest number of unique arguments being tagged for Nugunu (ISO
639-3: yas; 76) and the fewest number of unique predicates being tagged
for Rapanui (ISO 639-3: rap; 57). If we compare the number of unique
nouns and verbs, the fewest unique nouns are identified for ut-Hun
(ISO 639-3: uth; 59) and the fewest unique verbs for Bambalang (ISO
639-3: bmo; 54).

As noted above, this is significantly more than currently available
crosslinguistic tagged data in terms of language coverage, with the
closest similar corpus-based dataset being the UDT, in which only
106 of the 160 languages in the v2.14 release have more than 200 tagged
sentences. However, before we can use this dataset for investigation,
we need to be confident in the validity of the tags as well as the
features that can be extracted from this dataset. Accordingly, we
now turn to validation of the tagging accuracy of the IBM2 word alignment
POS-tag transfer method (§\ref{subsec:Validating-tag-accuracy}),
as well as how well extracted features correlate with existing classifications
in typological databases (§\ref{subsec:Testing-features-extracted}).

\subsection{Validating tag accuracy\label{subsec:Validating-tag-accuracy}}

One way to validate the automated tags provided by the IBM2 word
alignment method involves assessing against other automatic taggers
developed for high resource languages. Here we want to see that the
method used for tagging these languages aligns with existing language-specific
taggers at a level above chance. 

POS-taggers have been developed for high-resource languages, and the
current state of the art is represented by the SpaCy and Trankit
text processing libraries. These libraries have supervised POS-taggers
for 16 and 31 languages respectively (excluding languages with non-roman
scripts). For this comparison, each of the trained taggers in these
libraries was used to tag the reduced PBC verses for the respective
language and then the specific tags for arguments (NOUN, PROPN, PRON)
and predicates (VERB, AUX) were compared with the same verses tagged
via the word alignment method.\footnote{I should note here, first, that both libraries use a variation of
the universal tag set adopted by the UDT project and my methodology,
as outlined in footnote \ref{fn:For-the-purposes}. While there are
some differences, they apply to less common word classes and do not
affect arguments and predicates, which were the focus of this comparison.
Second, these taggers were not developed using Biblical texts, which
reduces the overall likelihood of correspondence, and makes the actual
result more striking.} Specifically, each word-alignment tagged verse was stripped of tags
and re-tagged by the respective libraries. Then the two sets of tagged
verses (between the library and the word-alignment method) were compared
and the number of matching arguments/predicates was divided by the
total number of arguments/predicates identified by the respective
library in a given sentence. This was then averaged across all verses
to provide an agreement percentage between the translation method
and the respective library for the argument and predicate tags.\footnote{See code at the linked repository for a more detailed breakdown of
correspondence between individual POS tags in the argument and predicate
classes.} Correspondence with argument and predicate tags between the word
alignment method and these pos-taggers was on average above 75\% for
both SpaCy (Table \ref{tab:SpaCy-correspondences-with}; arguments:
0.84; predicates: 0.76) and Trankit (Table \ref{tab:Trankit-correspondences-with};
arguments: 0.84; predicates: 0.77).

\begin{table}[h]
\noindent \begin{centering}
\begin{tabular}{llll}
\textbf{ISO 639-3} & \textbf{Language} & \textbf{Arguments} & \textbf{Predicates}\tabularnewline
cat & Catalan & 0.85 & 0.76\tabularnewline
hrv & Croatian & 0.76 & 0.56\tabularnewline
dan & Danish & 0.87 & 0.84\tabularnewline
nld & Dutch & 0.92 & 0.87\tabularnewline
fin & Finnish & 0.9 & 0.8\tabularnewline
fra & French & 0.85 & 0.67\tabularnewline
deu & German & 0.87 & 0.84\tabularnewline
ita & Italian & 0.87 & 0.8\tabularnewline
lit & Lithuanian & 0.82 & 0.63\tabularnewline
nob & Norwegian Bokmål & 0.91 & 0.86\tabularnewline
pol & Polish & 0.82 & 0.65\tabularnewline
por & Portuguese & 0.84 & 0.78\tabularnewline
ron & Romanian & 0.75 & 0.66\tabularnewline
slv & Slovenian & 0.6 & 0.72\tabularnewline
spa & Spanish & 0.88 & 0.88\tabularnewline
swe & Swedish & 0.9 & 0.79\tabularnewline
 & \textbf{Average} & \textbf{0.84} & \textbf{0.76}\tabularnewline
\end{tabular}
\par\end{centering}
\caption{SpaCy correspondences with the \emph{taggedPBC}\label{tab:SpaCy-correspondences-with}}

\end{table}

\begin{table}[h]
\noindent \begin{centering}
\begin{tabular}{llll}
\textbf{ISO 639-3} & \textbf{Language} & \textbf{Arguments} & \textbf{Predicates}\tabularnewline
afr & Afrikaans & 0.94 & 0.89\tabularnewline
eus & Basque & 0.69 & 0.78\tabularnewline
bul & Bulgarian & 0.81 & 0.7\tabularnewline
cat & Catalan & 0.84 & 0.88\tabularnewline
hrv & Croatian & 0.76 & 0.7\tabularnewline
ces & Czech & 0.77 & 0.71\tabularnewline
dan & Danish & 0.86 & 0.9\tabularnewline
nld & Dutch & 0.92 & 0.89\tabularnewline
fin & Finnish & 0.92 & 0.84\tabularnewline
fra & French & 0.87 & 0.82\tabularnewline
deu & German & 0.9 & 0.86\tabularnewline
hun & Hungarian & 0.91 & 0.68\tabularnewline
ind & Indonesian & 0.88 & 0.65\tabularnewline
gle & Irish & 0.82 & 0.46\tabularnewline
ita & Italian & 0.88 & 0.83\tabularnewline
lat & Latin & 0.86 & 0.73\tabularnewline
lav & Latvian & 0.91 & 0.82\tabularnewline
lit & Lithuanian & 0.88 & 0.76\tabularnewline
nob & Norwegian (Bokmål) & 0.91 & 0.89\tabularnewline
nno & Norwegian (Nynorsk) & 0.9 & 0.84\tabularnewline
pol & Polish & 0.83 & 0.65\tabularnewline
por & Portuguese & 0.86 & 0.8\tabularnewline
ron & Romanian & 0.76 & 0.8\tabularnewline
gla & Scottish Gaelic & 0.84 & 0.56\tabularnewline
srp & Serbian & 0.75 & 0.73\tabularnewline
slk & Slovak & 0.76 & 0.73\tabularnewline
slv & Slovenian & 0.6 & 0.75\tabularnewline
spa & Spanish & 0.89 & 0.9\tabularnewline
swe & Swedish & 0.89 & 0.9\tabularnewline
tur & Turkish & 0.85 & 0.6\tabularnewline
vie & Vietnamese & 0.86 & 0.69\tabularnewline
 & \textbf{Average:} & \textbf{0.84} & \textbf{0.77}\tabularnewline
\end{tabular}
\par\end{centering}
\caption{Trankit correspondences with the \emph{taggedPBC}\label{tab:Trankit-correspondences-with}}

\end{table}

This comparison informs us that the translation method agrees with
existing SOTA POS-taggers more than 75\% of the time, which suggests
that this methodology is relatively robust at transferring tags from
English to other languages. At least for tagging major word classes,
it aligns pretty well with automatic taggers individually developed
for specific languages. However, ``the proof is in the pudding'':
does the data in the \emph{taggedPBC} lend itself to actual correlations
with observed properties of language? To assess this, we now turn
to observations of word order that can be extracted from this dataset.

\subsection{Testing features extracted from the dataset: word order\label{subsec:Testing-features-extracted}}

By ``word order'' most researchers refer to the relative order
of referential arguments (i.e. elements referring to prototypical
agents/patients, often represented by nouns, proper nouns, and pronouns)
versus predicates (i.e. elements referring to prototypical actions
and states as represented by verbs and auxiliaries). This makes it
possible to talk about ``verb-initial'' vs ``verb-final'' or ``verb-medial''
languages, as well as grammatical constructs like Subject and Object
in relation to the position of the Verb. In the current investigation
I focus on intransitive word order, which is concerned with a predicate
that requires a single core argument, resulting in three possible
word orders: ``verb final'' (SV), ``verb initial'' (VS), and ``free''
(no dominant order). In order to identify these relations in corpora
I use the following procedure for each language in the \emph{taggedPBC}
corpus:
\begin{enumerate}
\item Filter the verses in the corpus, retaining only verses/sentences with
\textbf{both} an argument and a predicate.
\item In the resulting set, verses/sentences with an argument occurring
before a predicate are classed as “noun first”, while those with a
predicate occurring before an argument are classed as “verb first”.\footnote{Note that the terms ``noun first'' and ``verb first'' are simply
class/grouping labels (they could be termed ``X'' and ``Y'') used
to compute the N1 ratio. The terms do not constitute claims about
a particular sentence structure.}
\item This gives a count of ``noun first'' and ``verb first'' verses/sentences
for each language.
\item Turn this into the “N1 ratio” by dividing the total number of ``noun
first'' verses/sentences by the total number of ``verb first''
verses/sentences. 
\end{enumerate}
The resulting scalar value (the “N1 ratio”) allows us to make a plausible
quantitative inference based on observation of actual text data, rather
than making a categorical judgement on a language based on our knowledge
of it (which may be limited or expansive depending on the particular
language).

\begin{figure}[h]
\noindent \begin{centering}
\includegraphics[width=1.0\textwidth]{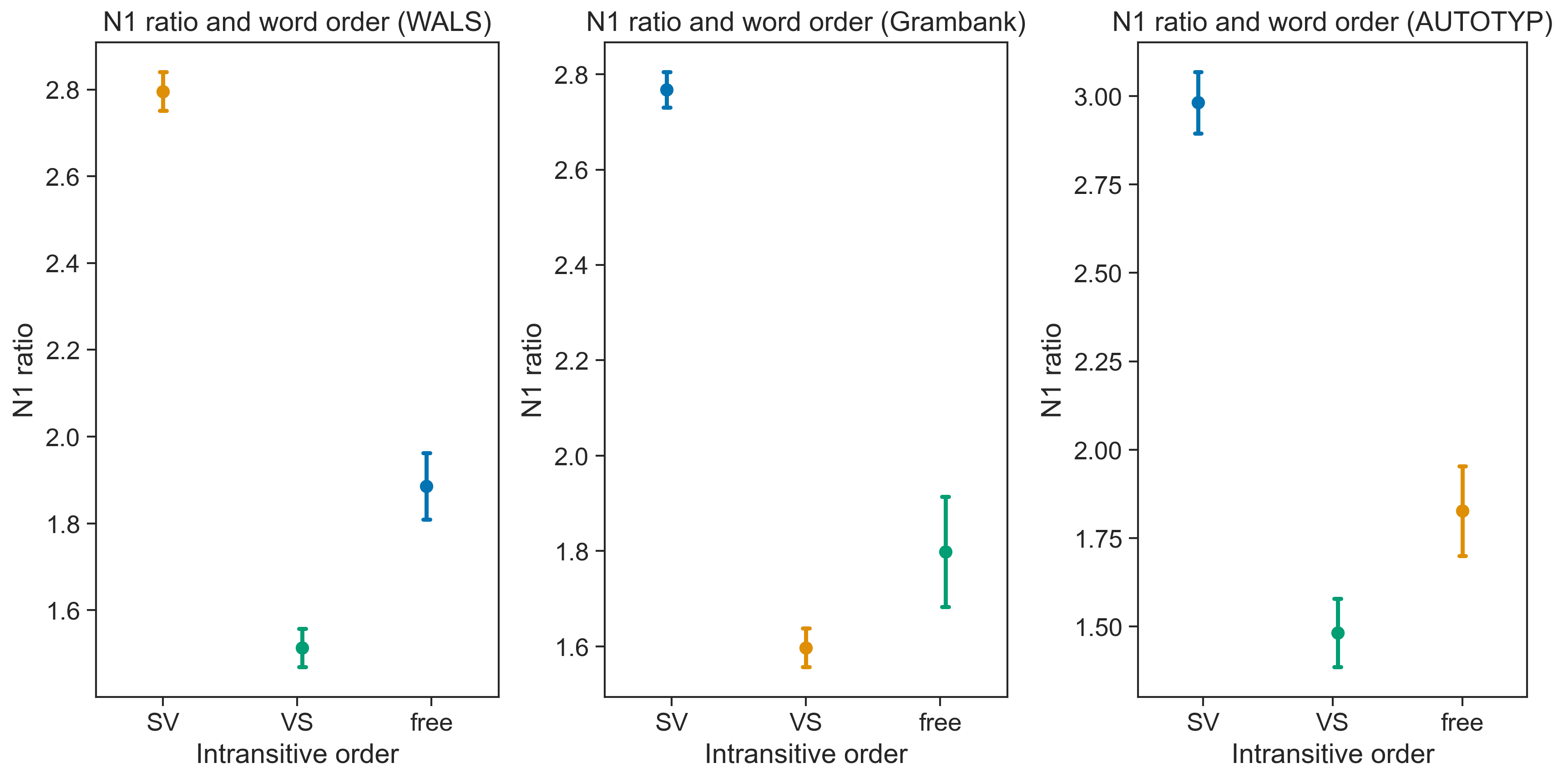}
\par\end{centering}
\caption{The N1 ratio in relation to word order in 3 typological databases\label{fig:The-N1-ratio}}

\end{figure}

To test the validity of this measure I compared the derived values
with known expert judgements of intransitive word order (SV, VS, or
``free'') in three typological databases (Figure \ref{fig:The-N1-ratio}):
the World Atlas of Language Structures (WALS; \cite{Dryer:2013ab}),
Grambank \cite{Skirgard:2023ab}, and the AUTOTYP Database \cite{Bickel:2023aa}.
For each of these databases, I conducted a one-way ANOVA with the
N1 ratio as the dependent variable, and the database category as the
grouping variable for all languages shared between the PBC and the
respective database (WALS: 646 languages; Grambank: 937 languages;
AUTOTYP: 203 languages).\footnote{Code to reproduce these results can be found at the linked repository.}
For each comparison, the N1 ratio significantly differentiated between
SV and VS languages (WALS mean diff. 1.443, p \textless{} 0.001; Grambank mean
diff. 1.285, p \textless{} 0.001; AUTOTYP mean diff. 1.544, p \textless{} 0.001), and
for WALS and Grambank it additionally differentiated between SV/VS
and free word orders,\footnote{For AUTOTYP the N1 ratio significantly differentiated between SV and
free word orders (p \textless{} 0.001) but was marginal for VS and free languages
(p = 0.054).} indicating that this ratio can be used as a proxy for word order,
at least for the PBC.

\begin{figure}[h]
\begin{centering}
\includegraphics[width=0.8\textwidth]{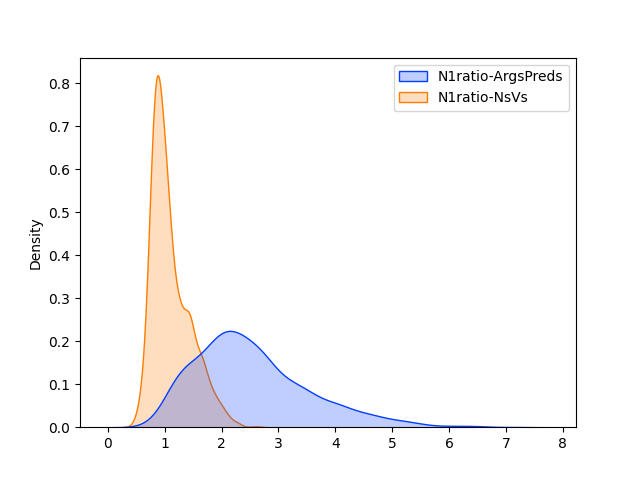}
\par\end{centering}
\caption{Distribution of the N1 ratio in the \emph{taggedPBC}\label{fig:Distrubution-of-the}}
\end{figure}

As an additional verification step of the N1 ratio as a proxy for
word order, I trained a machine learning classifier to predict word
order classifications for languages in the PBC for which word order
was unknown. Since the \emph{taggedPBC} data is normally distributed
(see Figure \ref{fig:Distrubution-of-the} and additional statistical
tests at the linked repository), I trained a Gaussian Naive Bayes
(GNB) classifier on the 1,152 languages with known word order (with
the N1 ratio as the feature), and then predicted on the remaining
790 languages with unknown word order. This resulted in 655 of the
languages being classified as ``SV'', and 135 being classified as
``VS''. Initial tests with other classifiers gave similar results,
though some (like the Decision Tree Classifier) identified up to 10
previously unclassified languages as ``free''. When literature searches
were conducted for these languages (see Table \ref{tab:Classifications-of-word}),
existing research identified all of the languages as having ``VS''
basic word order, validating the GNB classification.

\begin{table}[h]
\begin{centering}
\begin{tabular}{lllll}
\textbf{ISO 639-3} & \textbf{Language} & \textbf{GNB} & \textbf{DTC} & \textbf{Actual order}\tabularnewline
ifa & Amganad Ifugao & VS & free & VS\tabularnewline
prf & Paranan & VS & free & VS\tabularnewline
bpr & Koronadal Blaan & VS & free & VS\tabularnewline
cnl & Lalana Chinantec & VS & free & VS\tabularnewline
ayp & North Mesopotamian Arabic & VS & free & VS\tabularnewline
zty & Yatee Zapotec & VS & free & VS\tabularnewline
zar & Rincón Zapotec & VS & free & VS\tabularnewline
mih & Chayuco Mixtec & VS & free & VS\tabularnewline
spy & Sabaot & VS & free & VS\tabularnewline
zca & Coatecas Altas Zapotec & VS & free & VS\tabularnewline
\end{tabular}
\par\end{centering}
\caption{Classifications of word order by two different classifiers\label{tab:Classifications-of-word}}

\end{table}

This test of features indicates that the \emph{taggedPBC} is a valid
source of crosslinguistic comparative data. Despite potential inaccuracies
in tags, features extracted directly from these corpora correlate
well with expert tools and classifications, both in terms of tags
in individual languages and in terms of features such as word order.
This suggests some exciting new lines of research.

\section{Conclusions and future work}

The preceding sections have presented a new, massive dataset of parallel
text tagged for parts of speech in 1,942 languages via cross-lingual
POS-tag transfer, derived partially from the PBC \cite{Mayer:2014aa}
with additional data sourced from newly-accessible Bible translations.
The dataset is available for research on Github as the \emph{taggedPBC},
allowing for expansion/collaboration, and the tags for nouns and verbs
have been validated via comparison with taggers for high-resource
languages. Additionally, I have shown that the N1 ratio, a feature
extracted from this dataset on the basis of arguments and predicates,
differentiates between the intransitive word order (SV, VS, free)
of languages, allowing for the classification of previously unclassified
languages, and suggesting exciting new directions for research in
computational typology.

\subsection{Potential for expansion\label{subsec:Potential-for-expansion}}

The availability of the dataset via Github means that there are many
possibilities for expansion and refinement of the data. The current
release provides a baseline of POS tags via crosslingual transfer
using the IBM2 model, but other alignment and transfer methods could
be applied to increase accuracy for individual languages or groups
of languages. As noted, there are concerns with the applicability
of universal POS tags for specific languages and language families.
To address this, manually tagging portions of the individual corpora
could allow for the training of individual models on a per-language
basis that would improve accuracy. Some guidance along these lines
is presented at the linked repository to facilitate collaboration.
In each case, it would be important to validate such approaches, while
balancing accuracy with computational efficiency.

Individual datasets could also be expanded, either through tagging
of additional portions of the Bible present in the original PBC, or
by adding data from other (hand-tagged) corpora. While this would
affect the overall distribution of data, it might also benefit crosslinguistic
investigations in various ways. Multiple levels of annotation could
also be supported, either through automatic annotation via lexical
databases or manual annotation, provided that these are also validated/verified
in a principled manner.

\subsection{Potential research directions\label{subsec:Potential-research-directions}}

This dataset provides an important source of data for the growing
field of computational typology. As mentioned in footnote \ref{fn:It-should-be}
above, this field has primarily been focused on statistical methods
whereby different features of language (or of environment) are correlated
with determinations or classifications in typological databases (see
\cite{Jager:2025aa}). This kind of investigation is limited by
the availability of information on such features or classifications
for particular languages, and there may be some debate regarding how
to precisely categorize a language (with respect to word order, see
the large number of ``UNK'' {[}unknown{]} classifications for word
order in the Grambank database).

The dataset presented here, along with the example of extracting the
N1 ratio and observing its relation to word order, provides a potential
new direction for computational typology, whereby features can be
directly extracted from corpora. As with the example above, datasets
like the \emph{taggedPBC} allow researchers to identify corpus-based
features that correlate with (and/or differentiate between) expert
classifications in typological databases. These features can then
be used to classify previously unclassified languages, greatly expanding
the coverage of such databases, and directly addressing the issue
of linguistic diversity, as highlighted by \cite{Benitez-Burraco:2025aa}
and others. Further, corpus-derived measures allow for differing approaches
to linguistic questions, such as the gradient-based word order approach
proposed by \cite{Levshina:2023ab}. While the example above shows
that languages do not overlap very much in terms of the N1 ratio with
regard to basic intransitive word order classifications, the point
remains that no language is entirely SV or VS, supporting alternative
methods to answer related questions.

In general, if we consider POS tags to be one means of identifying
syntactic patterns, as suggested by the present investigation, there
should be a principled, crosslinguistic way to do so. Grouping POS
tags and/or observing their relative orders in sentences is one way
to compute measures from corpora. It logically follows, then, that
if such measures reliably correlate with other linguistic features
(as the N1 ratio does with intransitive word order in WALS/Grambank/AUTOTYP),
we should be able to use them as proxy measures for those features.
Having a tagged crosslinguistic dataset of this size is valuable for
answering multiple questions (such as whether a language has prepositions
or postpositions), and improving the scope and quality of the annotations
simply increases the kinds of questions that can be asked and answered.

Some questions that this dataset is uniquely able to support investigations
in are those related to word types crosslinguistically, whether with
regard to length, order, or other patterns. The fact that such questions
can be queried in relation to corpora additionally allows for known
effects of frequency on language \cite{Behrens:2016aa,Bentz:2017aa,Bybee:2006aa,Haspelmath:2021aa,Koplenig:2022aa}
to be accounted for. Other questions that could be asked, related
to dependencies and phonological properties, may require additional
annotation, whether automated, semi-automated, or manual. While improvements
can definitely be made to the overall annotations, the \emph{taggedPBC}
dataset provides a solid baseline to push computational typology forward
as a means of investigating (universal) properties of human language.

\bibliography{my-bibliog}

@misc{Zeman:2024ab,
	author = {Zeman, Daniel and Nivre, Joakim and Abrams, Mitchell and et al},
	copyright = {Licence Universal Dependencies v2.14},
	note = {{LINDAT}/{CLARIAH}-{CZ} digital library at the Institute of Formal and Applied Linguistics ({{\'U}FAL}), Faculty of Mathematics and Physics, Charles University},
	title = {Universal Dependencies 2.14},
	year = {2024}}

@article{Skirgard:2023ab,
	author = {Skirg{\aa}rd, Hedvig and Haynie, Hannah J. and Blasi, Dami{\'a}n E. and Hammarstr{\"o}m, Harald and et al},
	issue = {16},
	journal = {Science Advances},
	title = {Grambank Reveals Global Patterns in the Structural Diversity of the World's Languages},
	volume = {9},
	year = {2023}}

@article{Ostling:2016aa,
	author = {Robert {\"O}stling and J{\"o}rg Tiedemann},
	journal = {Prague Bulletin of Mathematical Linguistics},
	month = {October},
	pages = {125--146},
	title = {Efficient word alignment with {M}arkov {C}hain {M}onte {C}arlo},
	url = {http://ufal.mff.cuni.cz/pbml/106/art-ostling-tiedemann.pdf},
	volume = {106},
	year = {2016}}

@misc{Mayer:2014aa,
	address = {Reykjavik},
	author = {Mayer, Thomas and Michael Cysouw},
	booktitle = {Proceedings of The International Conference on Language Resources and Evaluation (LREC)},
	journal = {LREC},
	pages = {3158-3163},
	publisher = {LREC},
	title = {Creating a Massively Parallel {B}ible Corpus},
	year = {2014}}

@misc{Marchisio:2022aa,
	archiveprefix = {arXiv},
	author = {Kelly Marchisio and Conghao Xiong and Philipp Koehn},
	eprint = {2104.08721},
	primaryclass = {cs.CL},
	title = {{E}mbedding-{E}nhanced {G}iza++: Improving Alignment in Low- and High- Resource Scenarios Using Embedding Space Geometry},
	url = {https://arxiv.org/abs/2104.08721},
	year = {2022}}

@article{List:2022aa,
	author = {List, Johann-Mattis and Forkel, Robert and Greenhill, Simon J. and Rzymski, Christoph and Englisch, Johannes and Gray, Russell D.},
	date = {2022/06/16},
	doi = {10.1038/s41597-022-01432-0},
	id = {List2022},
	isbn = {2052-4463},
	journal = {Scientific Data},
	number = {1},
	pages = {316},
	title = {Lexibank, a public repository of standardized wordlists with computed phonological and lexical features},
	url = {https://doi.org/10.1038/s41597-022-01432-0},
	volume = {9},
	year = {2022}}

@article{Levshina:2023ab,
	author = {Natalia Levshina and Savithry Namboodiripad and Marc Allassonni{\`e}re-Tang and Mathew Kramer and Luigi Talamo and Annemarie Verkerk and Sasha Wilmoth and Gabriela Garrido Rodriguez and Timothy Michael Gupton and Evan Kidd and Zoey Liu and Chiara Naccarato and Rachel Nordlinger and Anastasia Panova and Natalia Stoynova},
	doi = {doi:10.1515/ling-2021-0098},
	journal = {Linguistics},
	lastchecked = {2025-04-08},
	number = {4},
	pages = {825--883},
	title = {Why we need a gradient approach to word order},
	url = {https://doi.org/10.1515/ling-2021-0098},
	volume = {61},
	year = {2023}}

@article{Koplenig:2022aa,
	author = {Koplenig, Alexander and Kupietz, Marc and Wolfer, Sascha},
	doi = {https://doi.org/10.1111/cogs.13090},
	eprint = {https://onlinelibrary.wiley.com/doi/pdf/10.1111/cogs.13090},
	journal = {Cognitive Science},
	number = {6},
	pages = {e13090},
	title = {Testing the Relationship between Word Length, Frequency, and Predictability Based on the {G}erman {R}eference {C}orpus},
	url = {https://onlinelibrary.wiley.com/doi/abs/10.1111/cogs.13090},
	volume = {46},
	year = {2022}}

@article{Jing:2021aa,
	author = {Yingqi Jing and Paul Widmer and Balthasar Bickel},
	journal = {Cognitive Science},
	title = {Word Order Variation is Partially Constrained by Syntactic Complexity},
	volume = {45},
	year = {2021}}

@article{Jager:2025aa,
	author = {Gerhard J{\"a}ger},
	journal = {arXiv},
	number = {cs.CL},
	title = {{C}omputational {T}ypology},
	url = {https://arxiv.org/abs/2504.15642},
	volume = {2504.15642},
	year = {2025}}

@misc{Imani:2022aa,
	address = {Abu Dhabi, United Arab Emirates},
	author = {Imani, Ayyoob and Severini, Silvia and Jalili Sabet, Masoud and Yvon, Fran{\c{c}}ois and Sch{\"u}tze, Hinrich},
	booktitle = {Proceedings of the 2022 Conference on Empirical Methods in Natural Language Processing},
	editor = {Goldberg, Yoav and Kozareva, Zornitsa and Zhang, Yue},
	month = {dec},
	pages = {1577--1589},
	publisher = {Association for Computational Linguistics},
	title = {Graph-Based Multilingual Label Propagation for Low-Resource Part-of-Speech Tagging},
	year = {2022}}

@misc{Imani:2022ab,
	archiveprefix = {arXiv},
	author = {Ayyoob Imani and L{\"u}tfi Kerem {\c S}enel and Masoud Jalili Sabet and Fran{\c c}ois Yvon and Hinrich Sch{\"u}tze},
	eprint = {2203.08654},
	primaryclass = {cs.CL},
	title = {Graph Neural Networks for Multiparallel Word Alignment},
	url = {https://arxiv.org/abs/2203.08654},
	year = {2022}}

@article{Henrich:2010aa,
	address = {New York, USA},
	author = {Henrich, Joseph and Heine, Steven J. and Norenzayan, Ara},
	doi = {10.1017/S0140525X0999152X},
	id = {cdi{\_}proquest{\_}miscellaneous{\_}954587569},
	isbn = {0140-525X},
	journal = {The Behavioral and Brain Sciences},
	journal1 = {Behav Brain Sci},
	n2 = {Behavioral scientists routinely publish broad claims about human psychology and behavior in the world's top journals based on samples drawn entirely from Western, Educated, Industrialized, Rich, and Democratic (WEIRD) societies. Researchers - often implicitly - assume that either there is little variation across human populations, or that these "standard subjects" are as representative of the species as any other population. Are these assumptions justified? Here, our review of the comparative database from across the behavioral sciences suggests both that there is substantial variability in experimental results across populations and that WEIRD subjects are particularly unusual compared with the rest of the species - frequent outliers. The domains reviewed include visual perception, fairness, cooperation, spatial reasoning, categorization and inferential induction, moral reasoning, reasoning styles, self-concepts and related motivations, and the heritability of IQ. The findings suggest that members of WEIRD societies, including young children, are among the least representative populations one could find for generalizing about humans. Many of these findings involve domains that are associated with fundamental aspects of psychology, motivation, and behavior - hence, there are no obvious a priori grounds for claiming that a particular behavioral phenomenon is universal based on sampling from a single subpopulation. Overall, these empirical patterns suggests that we need to be less cavalier in addressing questions of human nature on the basis of data drawn from this particularly thin, and rather unusual, slice of humanity. We close by proposing ways to structurally re-organize the behavioral sciences to best tackle these challenges.},
	number = {2-3},
	pages = {61--83},
	publisher = {Cambridge University Press},
	title = {The weirdest people in the world?},
	volume = {33},
	year = {2010}}

@article{Haspelmath:2021aa,
	author = {Haspelmath, Martin},
	doi = {10.1017/S0022226720000535},
	journal = {Journal of Linguistics},
	number = {3},
	pages = {605--633},
	title = {Explaining grammatical coding asymmetries: Form--frequency correspondences and predictability},
	volume = {57},
	year = {2021}}

@book{Hammarstrom:2024aa,
	address = {Leipzig},
	author = {Hammarstr{\"o}m, Harald and Forkel, Robert and Haspelmath, Martin and Bank, Sebastian},
	doi = {https://doi.org/10.5281/zenodo.14006617},
	note = {Accessed on 2025-04-03},
	publisher = {Max Planck Institute for Evolutionary Anthropology},
	title = {Glottolog 5.1},
	url = {http://glottolog.org},
	year = {2024}}

@article{Hahn:2020aa,
	author = {Michael Hahn and Dan Jurafsky and Richard Futrell},
	journal = {PNAS},
	number = {5},
	pages = {2347--2353},
	title = {Universals of word order reflect optimization of grammars for efficient communication},
	volume = {117},
	year = {2020}}

@article{Hahn:2022aa,
	author = {Michael Hahn and Yang Xu},
	journal = {PNAS},
	number = {24},
	title = {Crosslinguistic word order variation reflects evolutionary pressures of dependency and information locality},
	volume = {119},
	year = {2022}}

@book{Dryer:2013ab,
	editor = {Matthew S. Dryer and Martin Haspelmath},
	publisher = {Zenodo},
	title = {WALS Online (v2020.4)},
	type = {Data set},
	year = {2013}}

@book{Bybee:2006aa,
	address = {Oxford},
	author = {Bybee, Joan L.},
	publisher = {Oxford University Press},
	title = {Frequency of use and the organization of language},
	year = {2006}}

@article{Blasi:2022aa,
	author = {Dami{\'a}n E. Blasi and Joseph Henrich and Evangelia Adamou and David Kemmerer and Asifa Majid},
	doi = {https://doi.org/10.1016/j.tics.2022.09.015},
	issn = {1364-6613},
	journal = {Trends in Cognitive Sciences},
	number = {12},
	pages = {1153-1170},
	title = {Over-reliance on {E}nglish hinders cognitive science},
	url = {https://www.sciencedirect.com/science/article/pii/S1364661322002364},
	volume = {26},
	year = {2022}}

@misc{Bickel:2023aa,
	author = {Bickel, Balthasar and Nichols, Johanna and Zakharko, Taras and Witzlack-Makarevich, Alena and Hildebrandt, Kristine and Rie{\ss}ler, Michael and Bierkandt, Lennart and Z{\'u}{\~n}iga, Fernando and Lowe, John B},
	title = {The {AUTOTYP} database (v1.1.1)},
	year = {2023}}

@article{Bentz:2017aa,
	author = {Christian Bentz and Dimitrios Alikaniotis and Tanja Samard{\v z}i{\'c} and Paula Buttery},
	journal = {Journal of Quantitative Linguistics},
	number = {2-3},
	pages = {128-162},
	title = {Variation in Word Frequency Distributions: Definitions, Measures and Implications for a Corpus-Based Language Typology},
	volume = {24},
	year = {2017}}

@article{Benitez-Burraco:2025aa,
	author = {Ben{\'\i}tez-Burraco, Antonio},
	date = {2025/02/25},
	doi = {10.1007/s10339-025-01262-z},
	id = {Ben{\'\i}tez-Burraco2025},
	isbn = {1612-4790},
	journal = {Cognitive Processing},
	title = {The cognitive science of language diversity: achievements and challenges},
	url = {https://doi.org/10.1007/s10339-025-01262-z},
	year = {2025}}

@book{Behrens:2016aa,
	address = {Berlin},
	editor = {Heike Behrens and Stefan Pf{\"a}nder},
	publisher = {Walter de Gruyter},
	title = {Experience Counts: Frequency Effects in Language},
	year = {2016}}

@misc{Agic:2015aa,
	author = {Zeljko Agic and Dirk Hovy and Anders S{\o}gaard},
	booktitle = {Annual Meeting of the Association for Computational Linguistics},
	publisher = {Association for Computational Linguistics},
	title = {If all you have is a bit of the {B}ible: Learning {POS} taggers for truly low-resource languages},
	year = {2015}}

\end{document}